\documentclass[12pt]{article}

\usepackage[margin=1in]{geometry}

\usepackage{setspace}

\usepackage{graphicx}

\usepackage{enumitem}

\usepackage{microtype}
\usepackage{subfigure}
\usepackage{booktabs}
\usepackage{comment}
\usepackage{wrapfig}
\usepackage{arydshln}
\usepackage{enumitem}
\usepackage{multirow}
\usepackage{url}
\usepackage{natbib}
\usepackage{authblk}

\RequirePackage[colorlinks,citecolor=blue,linkcolor=blue,urlcolor=blue,pagebackref]{hyperref}

\usepackage{amsmath}
\usepackage{amssymb}
\usepackage{mathtools}
\usepackage{amsfonts}
\usepackage{bm}
\usepackage{bbm}
\usepackage{natbib}

\usepackage{algorithm}
\usepackage{algorithmic}
\def\eqref#1{equation~\ref{#1}}
\def\1{\bm{1}}

\def\bmX{{\bm{X}}}

\def\bmZ{{\bm{Z}}}

\DeclareMathAlphabet{\mathsfit}{\encodingdefault}{\sfdefault}{m}{sl}
\SetMathAlphabet{\mathsfit}{bold}{\encodingdefault}{\sfdefault}{bx}{n}

\def\calD{{\mathcal{D}}}

\def\calG{{\mathcal{G}}}

\newcommand{\cb}[1]{\left\{#1\right\}}

\usepackage{amsthm}

\theoremstyle{plain}

\usepackage[textsize=tiny]{todonotes}
\usepackage{multirow}
\usepackage{wrapfig}
\usepackage{subfigure}

\renewcommand{\eqref}[1]{(\ref{#1})}

\usepackage{xcolor}
\usepackage{ulem}
\newcount\Comments  % 0 suppresses notes to selves in text
\newcommand{\kibitz}[2]{\ifnum\Comments=1\textcolor{#1}{#2}\fi}

\usepackage{xurl}

\usepackage{listings,jvlisting}

\newcommand{\Vraw}{\mathcal{V}_{\mathrm{raw}}}
\newcommand{\Vcanon}{\mathcal{V}_{\mathrm{canon}}}

\usepackage[capitalize,noabbrev]{cleveref}

\allowdisplaybreaks

\title{Causality Elicitation from Large Language Models}

\author{Takashi Kameyama, Masahiro Kato\thanks{Email: \texttt{mkato-csecon@g.ecc.u-tokyo.ac.jp}}$\,$, Yasuko Hio, Yasushi Takano, Naoto Minakawa}

\affil{Mizuho-DL Financial Technology Co., Ltd.}

\date{\today}
\begin{document}

\maketitle 

\begin{abstract}
Large language models (LLMs) are trained on enormous amounts of data and encode knowledge in their parameters. We propose a pipeline to elicit causal relationships from LLMs. Specifically, (i) we sample many documents from LLMs on a given topic, (ii) we extract an event list from from each document, (iii) we group events that appear across documents into canonical events, (iv) we construct a binary indicator vector for each document over canonical events, and (v) we estimate candidate causal graphs using causal discovery methods. Our approach does not guarantee real-world causality. Rather, it provides a framework for presenting the set of causal hypotheses that LLMs can plausibly assume, as an inspectable set of variables and candidate graphs.
\end{abstract}

% ============================================================
\section{Introduction}
Causal analysis often begins by identifying key events and specifying hypothesized mechanisms that connect them.
The advent of large language models (LLMs) enables us to automate document summarization, event extraction, and the generation of causal narratives, thereby enriching such analyses. Moreover, the knowledge encoded in LLMs can be leveraged to propose causal hypotheses.
Importantly, our output graph is a ``hypothesis map'', not a validated real-world causal model.

This study develops a method to elicit causal relationships among events from LLMs.
We view each generated document as a topic-conditioned scenario sample and encode event co-occurrence in an incidence matrix.
For that purpose, we propose a general pipeline consisting of five steps:
\begin{itemize}
    \item \textbf{Step~(i):} generating documents from an LLM on a topic of interest;
    \item \textbf{Step~(ii):} extracting events that represent each document;
    \item \textbf{Step~(iii):} canonicalizing event mentions that appear across documents;
    \item \textbf{Step~(iv):} constructing a
    document--event incidence matrix from canonical events; and 
    \item \textbf{Step~(v):} conducting causal discovery using the resulting matrix.
\end{itemize}
The motivation is to establish a procedure for extracting causal hypotheses in a sophisticated, interpretable, and robust way within the framework of causal inference.
    
Steps (iii) and (iv) address the problem of variation in event names.
If events extracted by an LLM are treated as free-form text, the ``variable identity'' problem arises:
the same underlying event may appear under multiple names across documents.
For example, suppose we obtain the following extractions from a discussion of trade policy:
\begin{quote}
``From 2026 onward: reintroduce tariffs on major trading partners''\\
``From 2026 onward, maintain or expand tariffs''\\
``Protectionism intensifies, and trade friction increases''
\end{quote}
These statements may essentially refer to the same underlying event, which we might call ``tariff tightening.''
However, if we treat raw strings as distinct variables, they are counted as separate columns. This not only makes interpretation difficult but also undermines the stability of downstream
comparison, feature selection, and causal discovery.
As a foundation for making an LLM-based causality elicitation workflow viable, we explicitly incorporate:
\begin{itemize}
  \item \textbf{Event canonicalization (Step (iii)):} 
  defining a mapping from
  near-duplicate event mentions to canonical event names under a granularity policy chosen by the analyst; and
  \item \textbf{Matrix aggregation (Step (iv)):} applying this mapping to deterministically 
  merge columns and construct a stable document--event incidence matrix.
\end{itemize}

In Step (v), we apply causal discovery methods to the document--event incidence matrix to extract hypothesized causal links.
We then aim to externalize, as a graph, the causal structure that an LLM can plausibly posit.

Thus, our contribution is to propose an LLM-based causality elicitation pipeline, where (1) document generation, (2) event canonicalization, and (3) causal discovery play important roles. Our method can be used with existing documents, not only with documents generated by our pipeline. We note that the obtained graph may not reflect real-world causality and should be interpreted as causality encoded in the LLM. We expect that our method will be used for hypothesis formulation or causality-based summarization of documents.

Our method intersects several research areas, including event extraction, entity resolution and record linkage, semantic deduplication via embedding representations and clustering, text-as-data feature construction in the social sciences, and causal relation extraction and LLM-assisted causal modeling.
Existing studies treat identity through event coreference and record linkage, scale semantic similarity using embeddings, and treat causality through causal relation extraction and LLM-assisted causal modeling.
We connect these lines by placing 
\emph{event canonicalization (i.e., representation consistency of event variables)} as an explicit upstream module, constructing a stable document--event matrix, and then connecting it to causal discovery.
Through this connection, from variable identity to matrix construction to causal discovery, we aim to present causal hypotheses assumed by an LLM as inspectable and comparable candidate causal graphs.

\section{Related Work}
\label{sec:related}
Event extraction and cross-document event coreference are core NLP problems \citep{Cybulska2014usinga}.
Our event canonicalization is closely related in that it clusters event mentions, but 
our objective is different: to obtain
a stable variable set for downstream matrix construction.

Event unification is closely related to entity resolution and record linkage \citep{Fellegi1969atheory,Elmagarmid2007duplicaterecord,Christen2012datamatching}. 
Our embedding-first canonicalization instantiates this idea
for \emph{textual event names}, and the incremental LLM-assisted method maintains and updates a canonical registry.
Because identity granularity is analyst-dependent, we treat normalization and equivalence as a \emph{policy knob}.

Regarding clustering based on semantic similarity, prior work uses sentence embeddings and scalable retrieval to support semantic matching at scale \citep{Reimers2019sentencebert,Johnson2017billionscale}.
We use embeddings and clustering for merging, and use the LLM for \emph{naming} and \emph{adjudicating borderline cases}, to construct a human-readable canonical vocabulary that can be treated as variables downstream.

The text-as-data tradition studies document-to-feature transformations and stresses validation and interpretability \citep{Grimmer2013textas,Gentzkow2019textas}.
Our document--event binary incidence matrix uses event phrases as features, and canonicalization reduces redundancy and improves stability.

% \paragraph{Causal knowledge extraction.}
Causal relation extraction and causal knowledge base construction are widely studied \citep{Yang2022asurvey,Hassanzadeh2020causalknowledge,Radinsky2012learningcausality}.
Recent work also explores LLMs 
for causal discovery/modeling support, 
and causal reasoning \citep{Liu2025largelanguage,Ma2025causalinference}, and system-oriented work organizes LLM-generated causal fragments at scale \citep{Mahadevan2025largecausal}.
In contrast, we focus on \emph{variable fragmentation caused by surface variation in event expressions}, and connect canonicalization to causal discovery through a stable document--event matrix.

% \paragraph{Causal discovery.}
Causal inference in economics and policy evaluation is built on frameworks that address confounding and counterfactuals \citep{Pearl2009causalitymodels,Imbens2015causalinference}.
We treat the document--event matrix constructed from LLM-generated documents as observational data and apply multiple discovery algorithms to obtain candidate graphs as hypotheses about the dependencies assumed by the LLM.
Because the resulting graph does not guarantee real-world causality, any final causal claim requires external data, identification assumptions, and expert judgment \citep{Pearl2009causalitymodels}.

\section{Problem Formulation}
\label{sec:problem}
Assume we have $N$ documents, including ones generated from LLMs.
Let the list of event mentions extracted from document $i$ be:
\[
E_i = [e_{i,1}, e_{i,2}, \ldots, e_{i,|E_i|}],
\]
where each $e_{i,k}$ is a short text string describing an event such as a policy action, market movement, or scenario assumption.

Let $\Vraw$ be the set of unique raw event strings across all documents and let $M = |\Vraw|$.
Define a document--event binary matrix $\bmX\in\{0,1\}^{N\times M}$ by
\[
X_{i,m}=\mathbb{I}[\text{raw event } v_m \text{ appears in document } i].
\]
In practice, $M$ can be large and redundant due to surface-form variation.

\paragraph{Goal.}
Construct a mapping from raw event strings to canonical event labels:
\[
f:\Vraw \rightarrow \Vcanon,
\]
so that semantically equivalent (or policy-equivalent) raw events map to the same canonical label.
Using $f$, we aim to:
\begin{enumerate}[leftmargin=*]
  \item rewrite all event lists $\{E_i\}$ into canonicalized lists $\{E'_i\}$, and
  \item construct an aggregated matrix $\bmZ\in\{0,1\}^{N\times C}$, where $C=|\Vcanon|$.
\end{enumerate}

\section{Methods}
\label{sec:methods}
Our goal is to externalize causal hypotheses implicit in LLM-generated narratives as an interpretable set of variables (canonical events)
and candidate causal graphs.
At a high level, the pipeline is:
\begin{itemize}
    \item \textbf{Step~(i):} Document collection and generation,
    \item \textbf{Step~(ii):} Event extraction,
    \item \textbf{Step~(iii):} Event canonicalization (embedding + clustering + naming),
    \item \textbf{Step~(iv):} Matrix construction (binary encoding at the document level),
    \item \textbf{Step~(v):} Causal discovery.
\end{itemize}
In this section, we focus on canonicalization and matrix construction, and then connect them to causal discovery.

\subsection*{Step~(i): Topic-Conditioned Document Generation}
Given a topic $T$, we sample $N$ analytical documents from an LLM (e.g., $N=100$).
In a reference implementation, we prompt the model to behave as an analyst and to produce detailed narratives
grounded in concrete events and mechanisms.
Generated documents are stored per topic to enable reproducibility and comparison across runs. Below is a prompt example. 

\begin{quote}
system: You are an analyst who writes analytical documents on economics and international politics in English. Analyze the given topic in detail, grounding your discussion in concrete events and keeping in mind what causes what and what happens as a result. Note: it is currently January 2026.\\
user: Create an analytical document in English analyzing: "\{topic\}".
\end{quote}

\subsection*{Step~(ii): Event Extraction}
For each document $d_i$, we extract a list of event phrases using an LLM.
In practice, LLM outputs may vary in format (JSON arrays, Python-like lists, newline- or comma-separated bullets).
A robust extraction layer normalizes outputs into \texttt{list[str]} by attempting, in order,
(i) JSON parsing, (ii) Python literal parsing, and (iii) fallback splitting by newlines, commas, and bullet markers.
This yields event lists $\{E_i\}$. Below is a prompt example. 

\begin{quote}
system: You are a specialist at extracting "meaningful events" from English text.
% \medskip
skip

\#\# Purpose\\
From the input text, extract events such as incidents, judgments, policy changes, decisions made in meetings, changes in outlook, and recognition of risks.
% \medskip
skip

\#\# Extraction rules\\
- Each event should make clear what happened / was judged / was signaled.\\
- If possible, include timing (e.g., 'Dec 2025 meeting'), actor (e.g., government and firm), and outcome (e.g., stocks rose, interest rates fell, sentiment deteriorated) in the event name.\\
- Consolidate duplicates that describe the same content into a single item.
% \medskip
skip

\#\# Output (list format only)\\
Output a list consisting of English event names.\\
user: Extract important events from the following text.\\
\{text\}
\end{quote}

\subsection*{Step~(iii): Canonicalization}
\label{sec:embeddingfirst}
Pairwise LLM matching across a large vocabulary is expensive.
We therefore adopt an \textbf{embedding-first} strategy: compute embeddings for event mentions, cluster them, and then use an LLM to name
each cluster with a human-readable canonical label.

First, we flatten all event mentions and create an order-preserving unique list $\{u_1,\dots,u_M\}$ (e.g., via \texttt{unique\_preserve\_order}).
Embed each $u_m$ using an embedding model (e.g., \texttt{text-embedding-3-large}) and compute embeddings in batches for throughput.

\emergencystretch=2em
Second, given a maximum number of clusters $K_{\max}$ (e.g., 30), apply MiniBatchKMeans with $K=\min(K_{\max},M)$.
To approximate cosine similarity, we L2-normalize embeddings before applying Euclidean KMeans. Below is a prompt example. 

\begin{quote} 
system: You are an editor who writes a representative text (event) for a cluster of policy/economic scenario sentences. Return exactly one English text (event) that represents the given examples, 10 words or fewer. Constraint: avoid meaningless cluster names such as 'Other'.\\
user: Create exactly one English text (event) that represents the following examples, 10 words or fewer:\\
\{examples\}
\end{quote}

Third, for each cluster $c$, choose the top $m$ examples (e.g., 5) that are closest to the cluster centroid.
With normalized embeddings, inner products correspond to cosine similarity, enabling efficient ranking.

Finally, provide representative examples to an LLM and request exactly one canonical event phrase (at temperature 0).
This yields a human-readable canonical vocabulary.

\begin{algorithm}[t] 
\caption{Embedding-first event canonicalization}
\label{alg:embeddingfirst}
\begin{algorithmic}[1]
\REQUIRE Per-document event lists $\{E_i\}_{i=1}^N$, embedding model $\mathrm{Embed}$, max clusters $K_{\max}$, naming LLM $\mathrm{Name}$
\ENSURE Canonicalization map $f:\Vraw\to\Vcanon$, canonicalized event lists $\{E'_i\}$
\STATE Flatten all events and build an order-preserving unique list $U=[u_1,\dots,u_M]$
\STATE Compute embeddings $H \gets [\mathrm{Embed}(u_1),\dots,\mathrm{Embed}(u_M)]$ and L2-normalize each $H_m$
\STATE Apply MiniBatchKMeans with $K=\min(K_{\max},M)$ to obtain labels $\ell(m)\in\{1,\dots,K\}$
\FOR{$c=1$ \TO $K$}
  \STATE Select representative examples $R_c \subset \cb{u_m:\ell(m)=c}$ closest to the cluster centroid
  \STATE Generate a canonical label $\hat{u}_c \gets \mathrm{Name}(R_c)$
\ENDFOR
\STATE Define $f(u_m)\gets \hat{u}_{\ell(m)}$ for $m=1,\dots,M$
\FOR{$i=1$ \TO $N$}
  \STATE Replace each event $e$ in $E_i$ with $f(e)$ to obtain $E'_i$
\ENDFOR
\STATE \RETURN $f, \{E'_i\}$
\end{algorithmic}
\end{algorithm}

\begin{algorithm}[t] 
\caption{Incremental LLM-assisted event canonicalization (optional extension)}
\label{alg:incremental}
\begin{algorithmic}[1]
\REQUIRE Event lists $\{E_i\}_{i=1}^N$, candidate retrieval $\mathrm{Candidates}$, LLM matcher $\mathrm{LLMMatch}$
\ENSURE Canonicalized lists $\{E'_i\}$ and canonical registry $\mathcal{C}$
\STATE Initialize canonical registry $\mathcal{C}\gets\emptyset$
\FOR{$i=1$ \TO $N$}
  \STATE Initialize $E'_i \gets E_i$
  \FOR{$k=1$ \TO $|E_i|$}
    \STATE $t \gets \mathrm{clean}(E_i[k])$
    \IF{$t$ is empty}
      \STATE $E'_i[k]\gets t$
    \ELSE
      \STATE $S \gets \mathrm{Candidates}(\mathcal{C}, t)$
      \IF{$S=\emptyset$}
        \STATE Create a new canonical event $c$ with name $t$
        \STATE Register occurrence $(i,k)$ in $c$ and add $c$ to $\mathcal{C}$
        \STATE $E'_i[k]\gets t$
      \ELSE
        \STATE $(\texttt{match}, c, u) \gets \mathrm{LLMMatch}(t, S)$
        \IF{$\texttt{match}=\texttt{false}$}
          \STATE Create a new canonical event $c'$ with name $t$
          \STATE Register occurrence $(i,k)$ in $c'$ and add $c'$ to $\mathcal{C}$
          \STATE $E'_i[k]\gets t$
        \ELSE
          \STATE Register occurrence $(i,k)$ in $c$
          \IF{$u\neq c.\texttt{name}$}
            \STATE Update $c.\texttt{name}\gets u$
            \FOR{each occurrence $(j,\ell)$ of $c$}
              \STATE Rewrite $E'_j[\ell]\gets u$
            \ENDFOR
          \ELSE
            \STATE $E'_i[k]\gets c.\texttt{name}$
          \ENDIF
        \ENDIF
      \ENDIF
    \ENDIF
  \ENDFOR
\ENDFOR
\STATE \RETURN $\{E'_i\}, \mathcal{C}$
\end{algorithmic}
\end{algorithm}

\subsection*{Step~(iv): Binary Incidence Matrix Construction} 
Once the canonicalization map $f$ is available, the document--event matrix can be aggregated deterministically.
Group raw columns by their canonical labels and merge them by a logical OR (elementwise maximum):
\[
Z_{i,c} = \max_{m \in \mathcal{G}(c)} X_{i,m},
\]
where $\mathcal{G}(c)$ denotes the set of raw-event columns mapped to canonical event $c$.
The interpretation is that document $i$ contains canonical event $c$ if it contains any of its raw variants.

\subsection*{Step~(v): Causal Discovery}
Using the aggregated document--event matrix $\bmZ$, we estimate candidate causal graphs via causal discovery.
In a reference implementation, we apply:
\begin{itemize}
  \item \textbf{PC} (constraint-based) \citep{Spirtes2000causationprediction}: estimates a graph based on (conditional) independence tests (e.g., $\alpha=0.1$).
  \item \textbf{GES} (score-based) \citep{Chickering2002optimalstructure}: searches for a directed graph by score optimization.
  \item \textbf{LiNGAM} (functional model) \citep{Shimizu2006alinear}: estimates a causal ordering and an adjacency matrix via ICA-LiNGAM and DirectLiNGAM.
\end{itemize}
Estimated graphs are visualized (e.g., via Graphviz, via pydot) and stored per topic.

\begin{algorithm}[!ht] 
\caption{End-to-end pipeline: causality elicitation from an LLM}
\label{alg:pipeline}
\begin{algorithmic}[1]
\REQUIRE Topic $T$, number of documents $N$, max clusters $K_{\max}$
\ENSURE Candidate causal graphs $\{\hat{\calG}_{\mathrm{PC}},\hat{\calG}_{\mathrm{GES}},\hat{\calG}_{\mathrm{LiNGAM}}\}$
\STATE \textbf{Document generation:} sample $N$ documents $\calD=\{d_i\}$ from an LLM (or collect existing documents)
\STATE \textbf{Event extraction:} extract event lists $E_i$ from each $d_i$ using an LLM (with robust normalization)
\STATE \textbf{Canonicalization:} obtain $f$ and $\{E'_i\}$ via Algorithm~\ref{alg:embeddingfirst} (optionally refine via Algorithm~\ref{alg:incremental})
\STATE \textbf{Matrix construction:} build $\bmZ$ from $\{E'_i\}$; remove constant columns
\STATE \textbf{Causal discovery:} apply PC, GES, and LiNGAM to $\bmZ$ to obtain candidate graphs
\STATE \RETURN Candidate graphs
\end{algorithmic}
\end{algorithm}

\section{Empirical Analysis}
\label{sec:case}
This section provides two case studies.
In the first case study, we investigate the impact of President Trump's policies on Japan's economy after 2026.
In the second case study, we investigate the impact of U.S. investment in AI on gold prices.

\subsection{Case Study~I: President Trump's Policy Effect on Japanese Economy}

We illustrate the pipeline on a forward-looking policy topic with interacting mechanisms:

\begin{quote}
\emph{How President Trump's second-term deal-making approach could affect Japan's economy after 2026.}
\end{quote}
Using the analyst-style generation prompt (January 2026 perspective), we sample $N=100$ analytical documents from an LLM under this topic.
From each document, we extract a list of meaningful events, such as policy actions, decisions, shifts in outlook, market and FX states, and risk perceptions.
We then canonicalize event mentions using the embedding-first procedure (Section~\ref{sec:embeddingfirst}) with a maximum cluster count $K_{\max}=30$.
After LLM-based cluster naming and the removal of non-informative columns (all-0 or all-1), we retain $C=30$ canonical events and construct the document--event incidence matrix $\bmZ\in\cb{0,1}^{100 \times 30}$.

Table~\ref{tab:trump_japan_events_30} reports the canonical event set for this run.
The events span four recurring thematic blocks:
(i) \emph{tariff leverage and managed trade} (broad tariffs, Section~232, universal baseline tariffs, purchase-commitment deals),
(ii) \emph{technology and security restrictions} (export controls, technology and data-component controls, semiconductor alignment),
(iii) \emph{macro--financial and FX adjustment} (JPY volatility, intervention, currency-manipulation rhetoric, BoJ trade-offs), and
(iv) \emph{energy and terms-of-trade shocks} (sanctions and shipping costs, LNG agreements, oil-price hedging).

\begin{table}[t] 
\centering

\begin{tabular}{r p{0.93\linewidth}}
\hline
ID & Canonical event \\
\hline
1  & Monitor U.S. policy changes and economic actions through 2028. \\
2  & Japanese FDI in U.S. auto sector increases significantly. \\
3  & Japan negotiates tariff exemptions through investment and trade agreements. \\
4  & U.S. implements broad tariffs impacting global trade. \\
5  & Japan hedges against strong dollar and high oil prices. \\
6  & Japan's economic adjustments to U.S. policies post-2026. \\
7  & U.S. Treasury intensifies rhetoric on foreign currency manipulation concerns. \\
8  & U.S.-Japan LNG agreements reduce Japan's energy import costs. \\
9  & Stricter export controls on China affecting Japanese businesses. \\
10 & U.S.-China trade tensions escalate, impacting global supply chains. \\
11 & U.S. considers reviving Section~232 auto tariffs for negotiations. \\
12 & Stricter U.S. domestic-content rules for EVs and batteries anticipated. \\
13 & U.S. tariffs threaten Japan's auto industry and export margins. \\
14 & Increased localization in U.S. defense procurement and supply chains. \\
15 & Japan increases host-nation support, defense buys for U.S. concessions. \\
16 & Trump administration may use tariffs for international deal-making post-2026. \\
17 & Japan's trade terms and exports are impacted. \\
18 & Implementing stricter controls on technology and data components. \\
19 & U.S. adopts managed trade with strategic tariffs and deals. \\
20 & USMCA 2026 review may tighten auto rules of origin. \\
21 & Increased managed-trade agreements with purchase commitments and investment conditions expected. \\
22 & U.S. considers implementing a universal 10\% baseline import tariff. \\
23 & U.S. tightens export controls on China, focusing on technology. \\
24 & Japan aligns with U.S. on semiconductor export controls, investments. \\
25 & Japan intervenes to stabilize yen amid currency depreciation concerns. \\
26 & Sanctions and disruptions elevate oil prices and shipping costs. \\
27 & USD/JPY volatility expected due to rate differentials and political factors. \\
28 & Stricter Buy American rules for defense and federal procurement. \\
29 & Tariff impacts may weaken JPY and increase volatility. \\
30 & BoJ balancing inflation, yen stability, and global economic pressures. \\
\hline
\end{tabular}
\caption{Canonical event vocabulary for the Trump--Japan deal-making case study ($N=100$, $C=30$).}
\label{tab:trump_japan_events_30}
\end{table}

\begin{figure*}[t]
  \centering
  \includegraphics[width=0.98\textwidth]{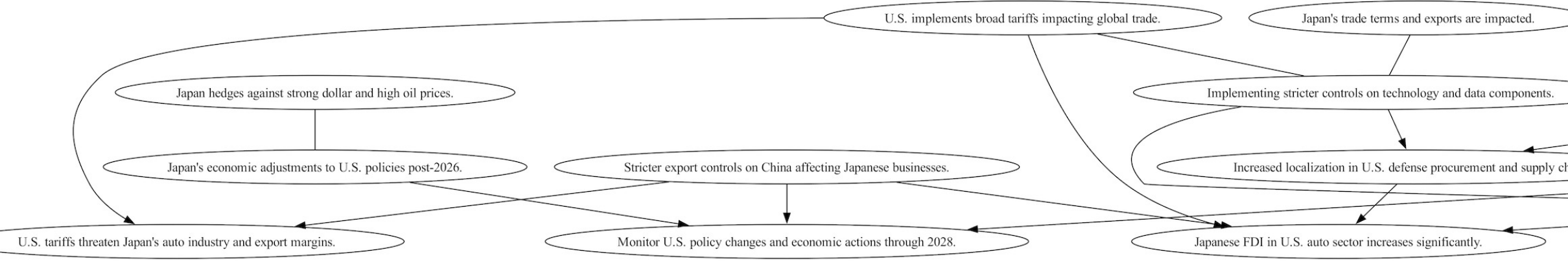}\par
  \vspace{0.8em} 
  \includegraphics[width=0.98\textwidth]{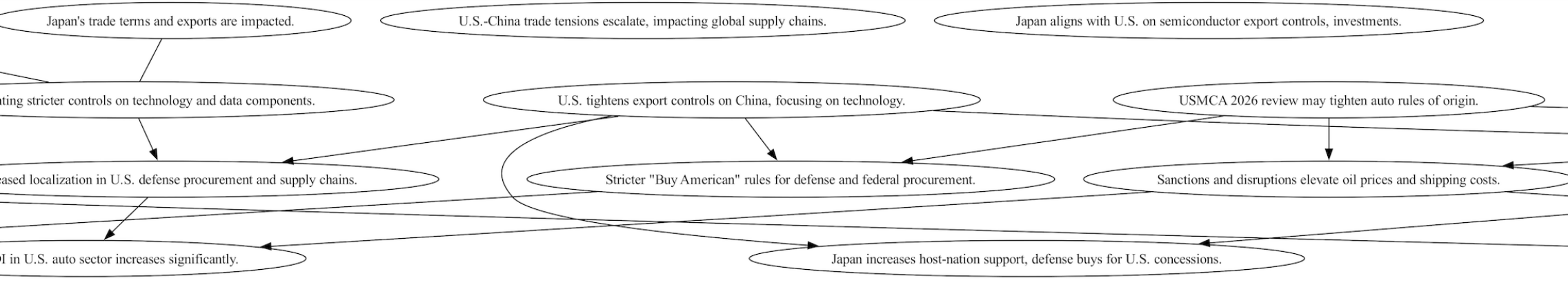}
  \caption{PC algorithm output for the Trump--Japan deal-making case study. ($N=100$, $C=30$). We split the original figure into two stacked panels for readability.}
  \label{fig:pc_trump_japan_30}
\end{figure*}

% \paragraph{Result of PC.}
First and foremost, we estimate a causal structure from $\bmZ$ using the PC procedure
(Figure~\ref{fig:pc_trump_japan_30}). The displayed PC subgraph highlights three mechanisms that repeatedly appear in the narratives:
\begin{itemize} 
    \item \textbf{(A) Technology restrictions $\rightarrow$ procurement localization $\rightarrow$ Japanese FDI.} \\
A directed chain links \emph{technology and data-component controls} (Event~18) and \emph{U.S. export controls focusing on technology} (Event~23) to
\emph{increased localization in U.S. defense procurement and supply chains} (Event~14), which points to
\emph{Japanese FDI in the U.S. auto sector} (Event~2).
In addition, \emph{broad U.S. tariffs} (Event~4) also point into Event~2.
This structure corresponds to a recurring narrative: U.S. policy tightens local-content and national-security constraints, and Japan responds by shifting investment and production footprints to the U.S. to reduce tariff exposure and satisfy localization requirements.

 \item \textbf{(B) Trade-rule tightening and procurement nationalism as a pressure bundle.} \\
\emph{USMCA 2026 rules-of-origin tightening} (Event~20) and \emph{U.S. export controls} (Event~23) point into
\emph{stricter Buy American rules for defense and federal procurement} (Event~28).
This reflects a bundled representation in the narrative space: trade-rule enforcement, export-control escalation, and procurement nationalism are repeatedly discussed as mutually reinforcing elements of post-2026 U.S. leverage.
A related node, \emph{sanctions and disruptions elevating oil prices and shipping costs} (Event~26), appears downstream in the displayed subgraph, capturing the frequent coupling of trade conflict with broader supply-chain and cost shocks.

 \item \textbf{(C) Japan-side response nodes and monitoring.} \\
Several Japan-side variables appear as sinks that collect multiple incoming edges.
First, \emph{host-nation support and defense purchases for U.S. concessions} (Event~15) receives incoming edges from multiple U.S.-side pressure variables in the displayed subgraph, consistent with narratives in which trade relief is negotiated alongside security burden-sharing.
Second, the meta-variable \emph{monitor U.S. policy changes and economic actions through 2028} (Event~1) is placed downstream of both
\emph{export controls affecting Japanese businesses} (Event~9) and \emph{Japan's economic adjustments} (Event~6).
While monitoring is not a primitive economic shock, its placement indicates that heightened policy uncertainty is represented through forward-looking attention language in the narratives.
Third, \emph{auto-industry margin stress from tariffs} (Event~13) is connected to the adjustment and monitoring block, underscoring that the auto sector acts as the central transmission margin in many sampled scenarios.
\end{itemize}

% \paragraph{GES and LiNGAM.}
In addition to PC, we also estimate GES and LiNGAM on the same incidence matrix (not shown in the main paper).
Because our inputs are binary incidence variables constructed from narrative samples, we treat edge orientation cautiously.
Qualitatively, the alternative estimators tend to preserve the same module-level structure observed in PC:
(i) a relocation and FDI adjustment module tied to trade pressure and localization requirements, and
(ii) a technology and security restriction bundle tied to procurement nationalism and broader supply-chain and cost shocks.
Differences primarily appear in how the methods orient edges among closely related policy instruments and among FX-related variables (Events~7, 25, 27, 29, 30), which are often highly correlated within narrative samples.

% \paragraph{Summary.}
In summary, this 30-event run illustrates that a finer canonical vocabulary can make the externalized hypothesis graph more interpretable.
Distinct U.S. leverage instruments (broad tariffs, export controls, Buy-American procurement, rules-of-origin tightening) map into distinct Japan-side adjustment channels (FDI relocation, tariff-exemption negotiation, defense and host-nation concessions), together with a monitoring sink that captures the forward-looking nature of post-2026 scenario narratives.
Overall, the PC provides a conservative summary of the conditional-dependence structure in the LLM-sampled scenario distribution, suitable as a front-end hypothesis map for domain-expert review. Indeed, the existence of analogous arguments 
\footnote{
\url{https://www.jri.co.jp/MediaLibrary/file/report/viewpoint/pdf/15710.pdf}; %, 
\par\url{https://www.jetro.go.jp/biz/areareports/special/2026/0101/0e88a5272a2374f5.html}} in the economic literature lends credibility to the three mechanisms proposed by the LLM.

\subsection{Case Study~II: U.S. Investment in AI and Gold Prices}
We provide a second case study in a macro--financial topic where multiple mechanisms plausibly coexist:

\begin{quote}
\emph{The impact of U.S. investment in AI on gold prices.}
\end{quote}

Using the same analyst-style document-generation prompt, we sample $N=100$ English analytical documents under this topic.
We extract per-document event lists, canonicalize event mentions using the embedding-first procedure (Section~\ref{sec:embeddingfirst}), and construct a document--event incidence matrix.
After canonical naming and the removal of non-informative columns, we obtain $C=20$ canonical events and an aggregated binary matrix $\bmZ\in\cb{0,1}^{N\times C}$.
Representative events include an AI capex surge, AI-driven growth narratives tied to USD strength and real yields, monitoring state variables such as TIPS and DXY, AI-chip export controls to China, Taiwan Strait tensions, and central-bank gold purchases.

\begin{figure*}[t]
\centering
\includegraphics[width=\textwidth]{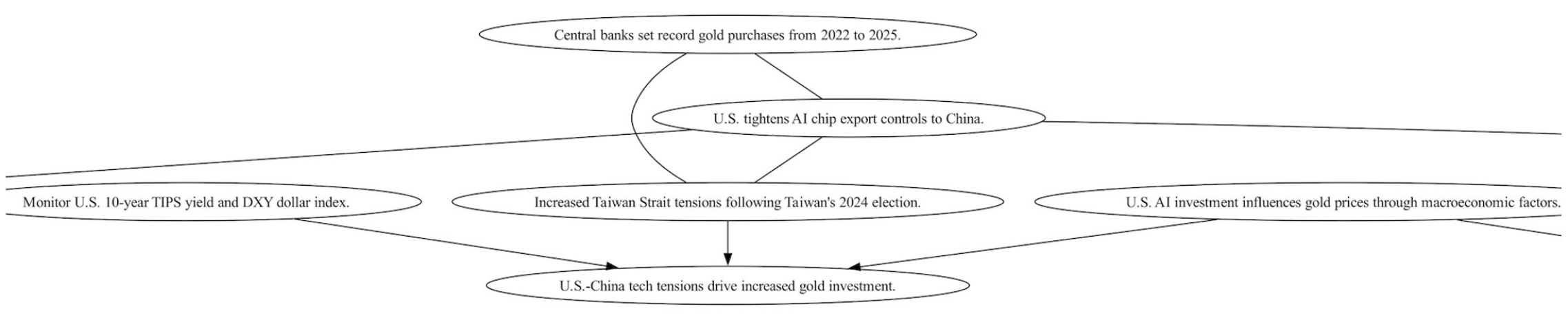}
\caption{Part of the PC result in the AI--gold case study ($N=100$, $C=20$).}
\label{fig:pc_ai_gold}
\end{figure*}

% \paragraph{Result of PC.}
First, we estimate causal structure from $\bmZ$ using the PC procedure (Figure~\ref{fig:pc_ai_gold}).
A key feature of the PC graph is that the canonical event describing U.S.--China tech tensions and increased gold investment acts as a hub.
It receives edges from Taiwan Strait tensions, a TIPS and DXY monitoring variable that summarizes macro--financial conditions invoked in the narratives, and an AI-to-macro mechanism that links AI investment to macro conditions.
In addition, the PC contains an undirected cluster among AI-chip export controls, Taiwan Strait tensions, and record central-bank purchases, indicating that these geopolitical and policy variables co-occur strongly in the scenario distribution, while their direction is not identified under the information available in $\bmZ$.

% \paragraph{Interpretation.}
As interpretation, overall, the PC suggests two channels in the LLM-sampled narratives.
The first is a macro--financial channel, in which AI investment is described as affecting growth and financial conditions, which are proxied by variables such as TIPS and DXY monitoring.
The second is a geopolitical and tech-rivalry channel, in which export controls and Taiwan-related risk are discussed alongside official-sector gold accumulation.
These channels converge on the gold-demand hub, providing a compact hypothesis map of how the LLM connects AI investment to gold prices.

% \paragraph{Summary.}
As a summary, even with a small canonical event set ($C=20$), the PC result externalizes an interpretable structure, separating a macro--financial pathway from a geopolitical and tech-rivalry pathway, and showing where they meet in the narratives.
We use this graph as a conservative summary of conditional dependence in the LLM-sampled scenario distribution, and treat it as a front-end object for domain-expert review.

\subsection{Limitations} % DiscussionがCase Study 5.1 & 5.2内でなされているので、Subsectionで良いかなと思料

% \section{Discussion and Limitations}
\label{sec:limitations}

First, canonicalization trades off false merges and missed merges. Whether time modifiers define identity or metadata is context-dependent and should be explicitly specified.

Second, extractions can mix events with states or mechanisms (e.g., ``uncertainty increases''). A type-classification layer is a natural extension.

Third, $\bmZ$ is binary, while PC tests and LiNGAM assumptions are typically developed for continuous data. Discrete tests/SEMs and methods robust to latent confounding (e.g., FCI) are important directions.

% \paragraph{Temporal ordering.}
Also, our incidence representation collapses within-document ordering, so edges primarily reflect conditional co-occurrence in the scenario distribution rather than temporal precedence.
When temporal precedence is available, it can be incorporated as directional constraints in causal discovery, e.g., disallowing edges that violate known time ordering.
%, making the resulting graphs more actionable as hypothesis maps for domain experts; temporal extensions are left for future work.

% \paragraph{Latent confounding and LLM sampling bias.}
In addition, LLM-generated documents can reflect omissions and prompt-dependent biases. Repeated-run stability checks and external validity assessments are needed, e.g., domain-expert review of high-confidence (run-stable) edges and validation against external datasets.

% \paragraph{Not causal verification.}
Finally, we do not verify causal truth. The graphs should be viewed as hypothesis spaces for experts to inspect, refute, and refine.

\section{Conclusion}
We proposed a pipeline to externalize causal structure from LLM outputs. To reduce surface-form variation, we canonicalize extracted events via embedding-based clustering with LLM-based naming and aggregate a document--event incidence matrix. We then apply PC, GES, and LiNGAM to obtain candidate causal graphs as hypothesis spaces. Future work includes methods for discrete data, latent confounding, stability analysis, external validity evaluation (e.g., domain-expert review), validation with external datasets (e.g., time-series or policy indicators), and temporal extensions.

\bibliographystyle{plainnat} 
\bibliography{arXiv3.bbl}

\end{document}